\documentclass[conference]{IEEEtran}
\IEEEoverridecommandlockouts
\usepackage{cite}
\usepackage{amsmath,amssymb,amsfonts}
\usepackage{algorithmic}
\usepackage{graphicx}
\usepackage{hyperref}
\newcommand{\loss}{\mathcal{L}}
\usepackage{caption}
\usepackage{textcomp}
\usepackage{xcolor}
\def\BibTeX{{\rm B\kern-.05em{\sc i\kern-.025em b}\kern-.08em
    T\kern-.1667em\lower.7ex\hbox{E}\kern-.125emX}}

\newcommand{\real}{\mathbb{R}}

\newcommand{\ex}{\mathbb{E}}

\newcommand{\bX}{\mathbf{X}}
\newcommand{\discri}{\mathcal{D}}
\newcommand{\G}{\mathcal{G}}

\newcommand{\bx}{\mathbf{x}}

\newcommand{\bz}{\mathbf{z}}

\usepackage{algorithmic, algorithm, setspace}

\newcommand{\yitian}[1]{\textcolor{black}{ #1}}

\makeatletter
\newcommand{\linebreakand}{%
 \end{@IEEEauthorhalign}
 \hfill\mbox{}\par
 \mbox{}\hfill\begin{@IEEEauthorhalign}
}
\makeatother

\usepackage{soul}
\usepackage{booktabs}

\begin{document}
\title{Contrastive Learning for Time Series\\on Dynamic Graphs
}

\author{\IEEEauthorblockN{Yitian Zhang}
\IEEEauthorblockA{\textit{Electrical and Computer Engineering}\\
\textit{McGill University}\\
Montreal, Canada\\
yitian.zhang@mail.mcgill.ca}
\and
\IEEEauthorblockN{Florence Regol}
\IEEEauthorblockA{\textit{Electrical and Computer Engineering}\\
\textit{McGill University}\\
Montreal, Canada\\
florence.robert-regol@mail.mcgill.ca}
\linebreakand
\IEEEauthorblockN{Antonios Valkanas}
\IEEEauthorblockA{\textit{Electrical and Computer Engineering}\\
\textit{McGill University}\\
Montreal, Canada\\
antonios.valkanas@mail.mcgill.ca}
\and
\IEEEauthorblockN{Mark Coates}
\IEEEauthorblockA{\textit{Electrical and Computer Engineering}\\
\textit{McGill University}\\
Montreal, Canada\\
mark.coates@mcgill.ca}}


\maketitle

\begin{abstract}
There have been several recent efforts towards developing representations for multivariate time-series in an unsupervised learning framework. Such representations can prove beneficial in tasks such as activity recognition, health monitoring, and anomaly detection. In this paper, we consider a setting where we observe time-series at each node in a dynamic graph. We propose a framework called \emph{GraphTNC} for unsupervised learning of joint representations of the graph and the time-series. Our approach employs a contrastive learning strategy. Based on an assumption that the time-series and graph evolution dynamics are piecewise smooth, we identify local windows of time where the signals exhibit approximate stationarity. We then train an encoding that allows the distribution of signals within a neighborhood to be distinguished from the distribution of non-neighboring signals. We first demonstrate the performance of our proposed framework using synthetic data, and subsequently we show that it can prove beneficial for the classification task with real-world datasets.
\end{abstract}

\section{Introduction}

Time series constitute a challenging data type for modeling, especially for supervised learning, due to their sparse labeling and complexity. To address this challenge, we can employ unsupervised methods to learn embeddings of the time series, and thereby extract informative low-dimensional representations. These general representations of the input data, derived without any need for labels, can be used for any downstream task.

In the last several years, self-supervised learning (SSL) has emerged as an effective strategy for learning representations. One form of SSL is contrastive learning, popularized by the SimCLR approach in~\cite{chen2020}.
One danger with self-supervised learning is {\em collapse}, when the model learns to output similar or even identical embeddings for all samples. Contrastive learning avoids collapse by identifying positive and negative training pairs. The embeddings of samples in a positive pair are encouraged to be similar, while those of samples in a negative pair are pushed apart. 

Several approaches have emerged for contrastive learning for time series. Contrastive Predictive Coding (CPC)~\cite{oord2019} is an effective strategy that first compresses high-dimensional data into a compact latent embedding space and then uses autoregressive models to predict the subsequent values of the signals. It uses predictive coding principles to train the encoder on a probabilistic contrastive loss.
Franceschi et al.\
employ a triplet loss in~\cite{franceschi2019}, which strives to ensure that a reference time series has a representation that is close to any one of its subseries (a positive sample) but far from negative series (chosen at random).
Temporal Neighborhood Coding (TNC) takes advantage of the local smoothness of the signals to learn generalizable representations for windows of a time series~\cite{tonekaboni2020}. This is achieved by ensuring that in the representation space, the distribution of signals that are close together in time is distinguishable from the distribution of signals that are far apart. TNC also takes into account the possibility that a pair of negative samples may also be similar.
    
Compared to contrastive learning, non-contrastive approaches are conceptually simple, and do not need a large batch size or a large memory bank to store negative samples. 
Notable approaches include \yitian{Bootstrap Your Own Latent} (BYOL)~\cite{grill2020} and \yitian{Simple Siamese} (SimSiam)~\cite{chen2021}. These methods train a student network to predict the representations of a teacher network. The weights of the latter are a moving average of the student's weights, or are shared with the student, but no gradient is backpropagated through the teacher. 
Recent efforts have explored the development of more effective loss terms. For example, \yitian{Variance-Invariance-Covariance Regularization (VICReg)} in~\cite{bardes2022} improves and builds upon the Barlow Twins loss of~\cite{zbontar2021}.

Some supervised learning frameworks have considered learning a graph structure to capture correlations in multivariate time series.
For example, in~\cite{hu2021}, Hu et al.\ propose EvoNet, which constructs a dynamic graph from time series data and can be used for event prediction. However, unsupervised representation learning of time-series on graphs remains underexplored in the literature.
    
In this paper, we propose a framework called~\emph{GraphTNC} for learning joint representations of the graph and the time-series. This procedure is designed for the setting where the underlying states of the signals and graph change over time. This model is also scalable to time-series with a static graph, where the graph input at each time-step are the same. We assess the quality of the learned representations on two datasets and show that the representations are general and transferable to downstream tasks such as classification.

Our contributions can be summarized as follows:
   \begin{itemize}
       \item We propose a novel encoder for learning representations of multivariate time series data on dynamic or static graphs, through a contrastive learning framework.
       \item We generalize non-contrastive learning methods from the computer vision domain to address non-stationary multivariate time series data.
   \end{itemize}
    
    

\section{Problem Setting}
\label{setting}

We consider the task of unsupervised learning of representations for time series on graphs. We denote a multivariate time series as $\bX\in\real^{N\times T}$ where $N$ is the number of univariate time series, and $T$ is the total length of the time series. A window of fixed length $w$ starting at time index $t$ is contained by the $[t, t+w]-$th columns of $\bX$: $\bX_{[t, t+w]}\in\real^{N\times w}$, and 
is denoted by $\mathbf{X}^t$. $w$ is assumed to be constant so we do not include it in the notation and only specify it in the text when essential for clarity. Associated with the multivariate time series, we also have dynamic graphs of $N$ nodes whose edges evolve alongside the time series. Each of the $N$ univariate time series is associated with one node in the graph. The edges between nodes are assumed to be indicative of the evolving correlation structure. 
Analogously to the windows for the time series, we denote \yitian{a window of dynamic graphs} by $\boldsymbol{\mathcal{G}}^t = [\G_{t},\dots,\G_{t+w}];\G_i = (\mathcal{V},\mathcal{E}_i), |\mathcal{V}| = N$ where each graph $\G_i$ is associated with the state of the graph at time $i$. Only the edge set $\mathcal{E}$ is indexed as the node set $\mathcal{V}$ stays constant.
The goal is to learn a representation $\bz^t\in\real^h$ of a time series window and its associated graphs $(\bX^{t}, \boldsymbol{\G}^t)$: $f^{enc}(\bX^{t}, \boldsymbol{\G}^t) =\bz^t$, where $h$ is the dimension of the joint representations. In this work, we subsequently use this representation to perform classification.

\section{Methodology}\label{metho}

We design an architecture that constructs a representation of a window of a multivariate time series and an associated sequence of graphs. The architecture is an encoder consisting of two modules. In the ensuing subsections we describe these modules and the loss function used to train the encoder. 

\subsection{Encoder  $f^{enc}(\bX^{t}, \boldsymbol{\G}^t)$}

Our encoding approach can be decomposed into two main building blocks : a) a \emph{static graph encoding module} that learns the state of the graph and its relationship with the multivariate time series; and b) a \emph{temporal module} which captures the dynamics of the data. 
\paragraph{Static graph encoding module} The purpose of this module is to learn the relationships between the node embeddings and the multivariate signal at a timestep $i$. To do so, we first need a representation of each individual node based on the state of the graph $\G_i$. This can be provided by any node embedding function $f^{\G}$ that takes a graph as input:
\begin{align}
\mathbf{H}_i & = f^{\G}(\G_i),\quad\mathbf{H}_i\in \real^{N\times k } ,
\end{align}
where $k$ is the dimension of the output node embeddings.\\
Next, we concatenate $\mathbf{H}_i$ with the time series of this timestep denoted by $\bx_i\in\real^N$ and pass it through a neural network to obtain \yitian{$\mathbf{e}_i$}, the final representation of the graph-signal interaction at timestep $i$: 
\begin{align}
\mathbf{e}_i &=  NN^1([\mathrm{vec}(\mathbf{H}_i)||\bx_i]),\quad\mathbf{e}_i\in\real^{d},
\end{align}
where $d$ is the dimension of the graph-signal interaction representation, \yitian{ $\mathrm{vec}(\cdot)$ is an operator that stacks the columns of a matrix, and $[\cdot||\cdot]$ denotes the concatenation of two vectors ($\mathrm{vec} :\real^{a\times b}\to\real^{ab}$, $[\cdot||\cdot] :\real^{a}||\real^{b} \to \real^{a+b}$).}

\paragraph{Temporal Module} 

We use a temporal-based neural network $f^{temp}$ to capture the dynamic nature of the data $(\bX^{t}, \boldsymbol{\G}^t)$ and to obtain the final representation $\bz^t$. The network $f^{temp}$ outputs the hidden state of the next timestep $\mathbf{s}_{i+1}\in\real^{s}$ based on the current hidden state $\mathbf{s}_i\in\real^{s}$ and on the input at time $i$. In our framework, the input is constructed of the signal $\bx_i$ concatenated with the processed graph-signal interaction \yitian{$\mathbf{e}_i$}. The final representation $\bz_t$ is obtained by passing the last hidden state of the window $\mathbf{s}_{w}$ through a neural network:
\begin{align}
  \mathbf{s}^t_{i+1} =& f^{temp}([\bx_{t+i}||\mathbf{s}^t_i ||\mathbf{e}_{t+i}])\\
\bz^t =& NN^2(\mathbf{s}^t_{w}) \label{eqn:e2}
\end{align}
We use a 1-layer graph convolution as $f^{\G}$ and a 1-layer bi-directional \yitian{Gated Recurrent Unit} (GRU) as $f^{temp}$. Both $NN^1$ and $NN^2$ are 1-layer feed-forward neural networks (FNN).

\subsection{Loss function}
\yitian{We define a discriminator $D(\bz^t,\bz)$. The objective function is to make the probability likelihood estimation of the discriminator to be close to 1 if $\bz$ and $\bz^t$ are representations of neighboring windows, and close to 0 otherwise.} Following~\cite{tonekaboni2020}, we view windows that are close in time as neighboring windows, and use the Augmented Dickey-Fuller (ADF) statistical test to find the neighborhood range. The neighbourhood $\mathcal{N}^t$ is selected based only on time series, since the underlying states of graphs and signals are assumed to evolve together. 

The loss function is defined as:
\begin{align}
\label{loss}
   \loss(\bX^t,\boldsymbol{\G}^t) &= -\Big[ \ex_{(\bX^{l}, \boldsymbol{\G}^l)\sim\mathcal{N}^t}\big[\log\discri (\bz^t,\bz^l)\big]\nonumber +\\ \ex_{(\bX^{k}, \boldsymbol{\G}^k)\sim\bar{\mathcal{N}}^t} &\big[(1-m)\log(1-\discri (\bz^t,\bz^k)) + m\log\discri  (\bz^t,\bz^k) \big] \Big]
\end{align}
where $m$ is the probability of sampling a positive window from the non-neighboring region $\bar{\mathcal{N}}^t$. By optimizing this function, representations
$\bz^l = f^{enc}(\bX^{l},\boldsymbol{\G}^l) $ of samples from a neighborhood $(\bX^{l}, ~\boldsymbol{\G}^l)\in\mathcal{N}^t$, can be distinguished from representations
$\bz^k = f^{enc}(\bX^{k}, \boldsymbol{\G}^k)$ of samples from outside the neighborhood.

\section{Experiments}

In the experiments, we evaluate the performance of our proposed model on a synthetic dataset in a controlled setting, and on a real-world dataset. Both datasets have underlying states that change over time. Therefore, a state is associated with each time window \yitian{$(\bX^t,~\boldsymbol{\G}^t)$}. The performance of the learned representations $\bz$ is evaluated by the downstream classification task, where the states are the classification targets. Below we describe the datasets, experiment setup and results in details.

\subsection{Datasets}
\subsubsection{Synthetic data} 
The synthetic dataset contains a multivariate time series influenced by a dynamic graph, that is also synthetically generated. The generation of both the time series and the graphs is driven by an underlying state of the time series that is modeled by a Hidden Markov Model (HMM).
In each state, the time series are generated from a different generative process including Nonlinear Auto-regressive Moving Average models with different sets of parameters and Gaussian Processes with different kernel functions, in a similar fashion to that employed in the synthetic experiment in~\cite{tonekaboni2020}. The features at each time step are concatenated in a vector $\mathbf{f}_t\in\real^{N}$.\\
For the graph structure, each state has a different initial random graph $\G^s_0$ generated from an Erdős–Rényi model with probability $p^s$ of an edge between nodes, where $s$ is the state number. With a probability $q^s = p^s/10(1-p^s)$, $\G^s_{t-1}$ adds new edges or removes existing ones to generate $\G^s_t$.\\
The time series data is generated by:
\begin{align}
\label{generate}
    \mathbf{x}_{t+1} = r  A_t\mathbf{f}_t + (1-r)\mathbf{f}_t,\quad 0\leq r\leq 1 ,
\end{align}
where $A_t$ is the adjacency matrix of $\G_t$, and $r$ weights how much influence the graph has on the time series.

\subsubsection{EEG}
EEG signals are recorded from probes connected to brains of human subjects. 
The dataset, which originates from an online data science competition\footnote{\href{https://www.kaggle.com/c/grasp-and-lift-eeg-detection/data}{https://www.kaggle.com/c/grasp-and-lift-eeg-detection/data}}, contains 32 channels of EEG recordings of subjects performing hand grasping and lifting actions. 
Each hand action is divided into 7 states that include: the initial movement, first touch of the object to be lifted, start of loading phase, hand lift off, change of hands, release and no action. 
The graph structure for this dataset encodes the spatial relationship between the 32 electrode locations. 
The dataset provides a map of the physical locations of the electrode probes with respect to the brain. These probes are arranged in a grid. We define a graph where each node represents an electrode probe and an edge connects two probes if they are direct neighbors on the grid. Since the graph is static, it is repeated at each time step to fit our model. We extract 100 signals of length 60 timesteps to train our model.

\subsection{Experiments Setup}
\subsubsection{GraphTNC vs baseline TNC}

In this experiment, we compare the classification performance of the representations learned from time series with and without considering the graph. For fair comparison, we use the same encoder proposed for the baseline TNC~\cite{tonekaboni2020} as $f^{temp}$, which is a 1-layer bi-directional GRU. In our architecture, we add a graph encoding module $f^{\G}$ (a 1-layer graph convolution) before the temporal module, such that the input of $f^{temp}$ is the combined information of the graph and signal. We also followed the setting $m=0.05$ in~\eqref{loss} from~\cite{tonekaboni2020} which is the probability of positive  window in non-neighboring region. Detailed hyperparameters can be seen in Table~\ref{para}. 

\begin{figure}
    \vspace{-0.3in}
   \centering
   \includegraphics[width=0.5\textwidth]{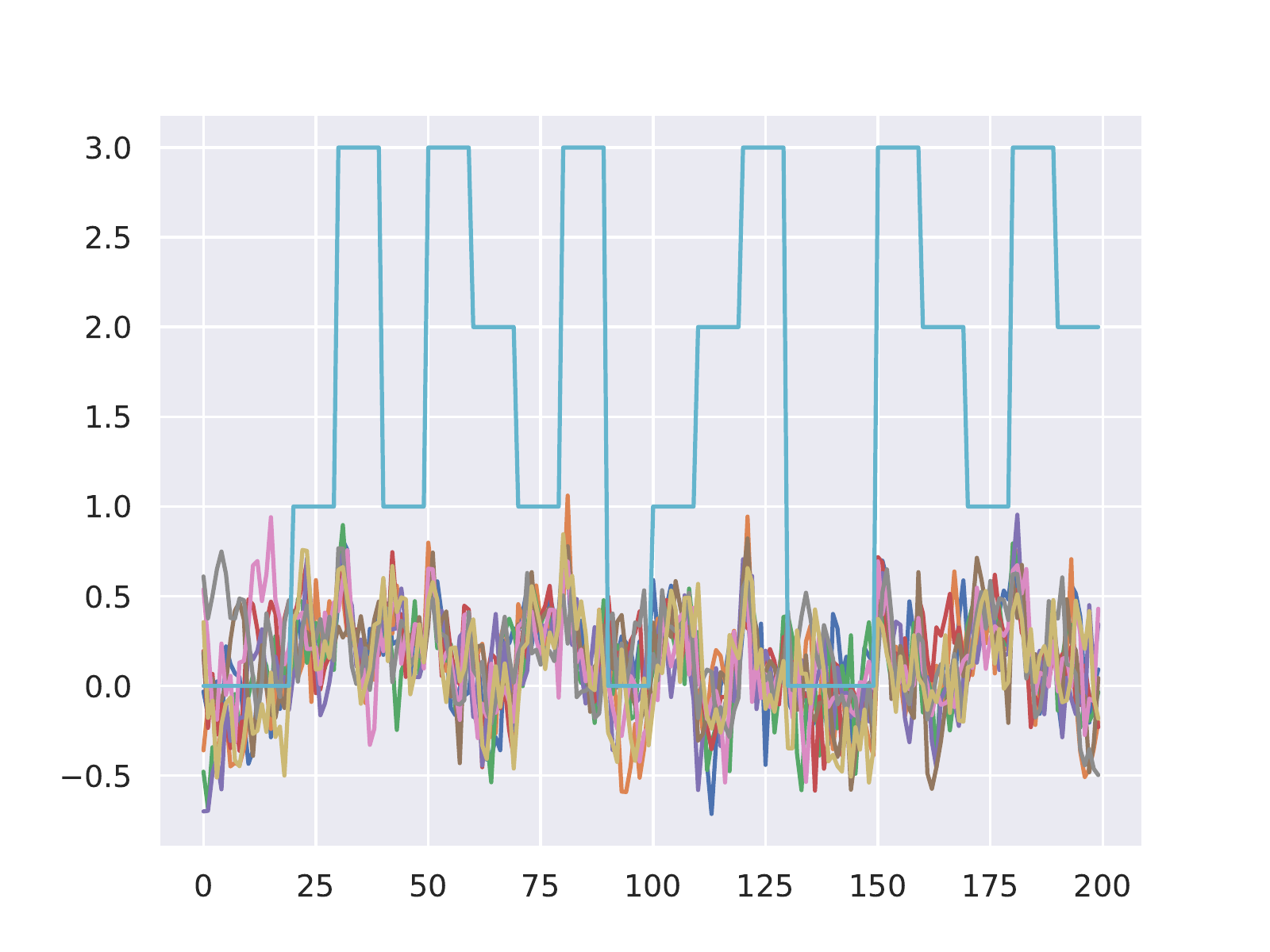}
   \caption{Synthetic dataset: Light blue discrete signal represents underlying state. Continuous signals represent the $N$ features.}
   \label{fig:synthetic_data}
\end{figure}

\begin{table}[tbp]
\small
\caption{Hyper-parameters of GraphTNC}
\begin{center}
\begin{tabular}{lll}
\toprule
 &\textbf{Synthetic} &\textbf{EEG}\\
\midrule
\textbf{Graph node (feature) number} $N$ & 10 & 32\\
\textbf{Graph encoding size} $k$ & 4 & 4\\
\textbf{Graph-signal interaction size} $d$ & 8 & 32\\
\textbf{GRU input size} ($N+d$)& 18  & 64\\
\textbf{GRU hidden size} & 64 & 64\\
\textbf{Joint representations size} $h$ & 8 & 32\\
\bottomrule
\end{tabular}
\label{para}
\end{center}
\end{table}

\begin{figure*}
   \centering
   \vspace{-0.1in}
   \includegraphics[trim={2.5cm 0cm 4cm 2cm},clip, width = 0.79\textwidth]{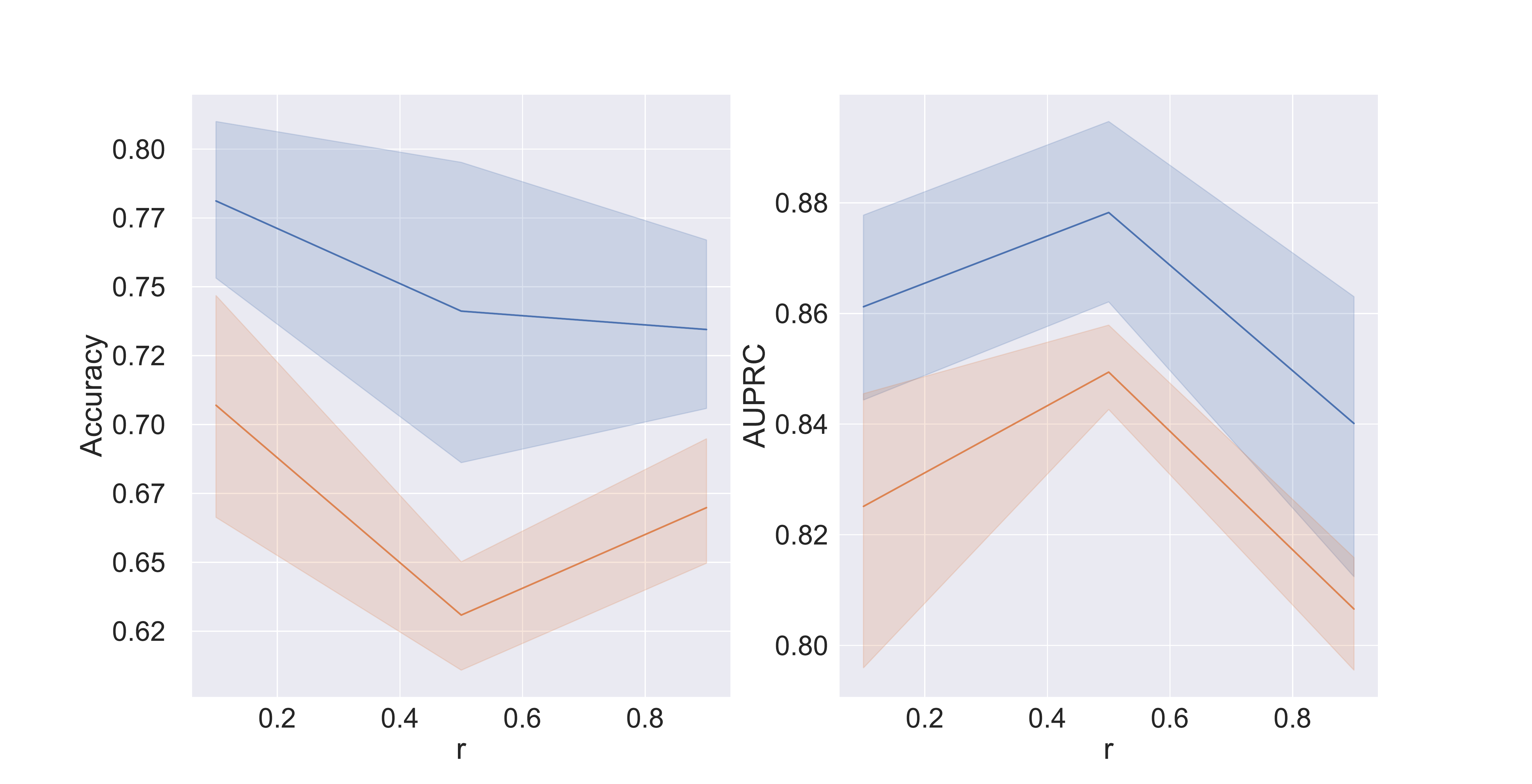}
   \caption{Results (with 95\% confidence intervals shaded) of synthetic dataset experiment with various $r$ values from eq.\eqref{generate}. Left: Accuracy results. Right: Area under precision recall curve (AUPRC). Our GraphTNC is shown in blue, baseline is in orange. }
   \label{fig:synthetic_results}
\end{figure*}

For training, we use the Adam optimizer with a learning rate of $1e{-}3$ and a weight decay of $1e{-}5$ and train for $100$ epochs with early stop for both datasets. As stated in Section~\ref{metho}, although the discriminator and encoder are learned together during the training phase, only the encoder is required during inference. The window size $w$ is selected through experiments such that it is long enough to contain information of underlying state but not too long to span over multiple states, following the same rationale in TNC~\cite{tonekaboni2020}.

To evaluate the quality of the representations, we use classification as a downstream task. \yitian{ The classifier is a 1-layer FNN on top of the frozen representation} with $h$ as input dimension and $S$ as output dimension, trained on cross-entropy loss. \yitian{We use the simple structure to reduce the impact of the classifier on the final results.} The performance is reported as the prediction accuracy and the area under the precision-recall curve (AUPRC) score since AUPRC is a more accurate reflection of model performance for imbalanced data.

\subsubsection{GraphTNC vs non-contrastive learning}
Throughout this experiment, we retain our proposed encoder $f^{enc}(\bX^{t}, \boldsymbol{\G}^t) $, and compare the performance of the GraphTNC contrastive learning approach with two non-contrastive learning methods, BYOL~\cite{grill2020} and SimSiam~\cite{chen2021}. BYOL has an asymmetric architecture where the weights $\theta_m$ of one encoder are an exponential moving average of the other encoder’s weights $\theta$. A predictor $g$ with weights $\phi$ is used in the branch with learnable weights. SimSiam uses a predictor on one branch and a stop-gradient operation on the other. In the original papers~\cite{grill2020,chen2021}, the inputs to the two encoders are the original image and an augmentation. 

To generalize these methods to our setting, we feed $(\bX^{t}, ~\boldsymbol{\G}^t)$ and $(\bX^{l}, ~\boldsymbol{\G}^l)\in\mathcal{N}^t$
as positive pairs to the student and teacher networks. Non-neighborhood samples are not required. 
For the projector and predictor for both BYOL and SimSiam, we use a 2-layer FNN with size 128-128.
We then compare the performance of all unsupervised approaches to a supervised model where the classifier and the encoder are trained end-to-end. For the supervised setting, the architectures of the encoder and the classifier are the same as for the unsupervised models. We evaluate the performance of the representations learned from different methods via the state classification accuracy and AUPRC. 

\begin{table}[tbp]
\small
\vspace{0.05in}
\caption{Classification results. The asterisk represents statistically significant result using paired Wilcoxon test ($p < 0.05$).}
\begin{center}
\begin{tabular}{lcccc}
\toprule
\textbf{}&\multicolumn{2}{c}{\textbf{GraphTNC (ours)}}&\multicolumn{2}{c}{\textbf{TNC}}\\
\midrule
\textbf{Dataset} &\textbf{\textit{AUPRC}}&\textbf{\textit{Accuracy}}&\textbf{\textit{AUPRC}}&\textbf{\textit{Accuracy}}\\\hline

$r=0.1$ &\textbf{0.86$\pm$0.03*} &\textbf{0.78$\pm$0.06*}  & 0.83$\pm$0.04 & 0.71$\pm$0.06\\
$r=0.5$ &\textbf{0.88$\pm$0.03*} &\textbf{0.74$\pm$0.09*}  & 0.85$\pm$0.01 & 0.63$\pm$0.03\\
$r=0.9$ &\textbf{0.84$\pm$0.04*} &\textbf{0.73$\pm$0.05*}  & 0.81$\pm$0.04 & 0.61$\pm$0.02\\
\hline
EEG  & 0.54$\pm$0.02&\textbf{0.92$\pm$0.02} & 0.54$\pm$0.02 &0.90$\pm$0.03\\
\bottomrule
\end{tabular}
\label{tsgraph}
\end{center}
\vspace{-0.05in}
\end{table}

\begin{table*}[tbp]
\small
\caption{Classification performance of joint representations of TS and Graph learned from different frameworks.}
\begin{center}
\begin{tabular}{lcccccc}
\toprule
\textbf{}&\multicolumn{3}{c}{\textbf{Synthetic $r=0.1$}}&\multicolumn{3}{c}{\textbf{EEG}}\\
\midrule

\textbf{} &\textbf{\textit{AUPRC}}&\textbf{\textit{Accuracy}}& \textit{Params} &\textbf{\textit{AUPRC}}&\textbf{\textit{Accuracy}}& \textit{Params}\\
\midrule
\textbf{GraphTNC (Ours)}&\textbf{0.86$\pm$0.03} &\textbf{0.78$\pm$0.06}  &34k &0.54$\pm$0.02&\textbf{0.92$\pm$0.02}&59k\\



\textbf{BYOL} & 0.73$\pm$0.10 & 0.56$\pm$0.10 & 67k &\textbf{0.55$\pm$0.02} &0.91$\pm$0.03 &92k\\

\textbf{SimSiam} & 0.74$\pm$0.10&0.57$\pm$0.10 &67k &\textbf{0.55$\pm$0.03} &0.90$\pm$0.03 & 92k\\

\hline
\textbf{Supervised} &\textbf{0.95$\pm$0.04} &\textbf{0.83$\pm$0.09} &34k &\textbf{0.57$\pm$0.03} &\textbf{0.92$\pm$0.01} &60k\\
\bottomrule

\end{tabular}
\label{non-con}
\end{center}
\end{table*}

\subsection{Experiments results and discussion}

\subsubsection{GraphTNC vs baseline TNC}

Table~\ref{tsgraph} presents the state classification results. An asterisk indicates a statistically significant difference at the $5\%$ level between the GraphTNC and the baseline for a Wilcoxon signed-rank test.
First, we can observe that our encoder, which learns joint representations of the time series and of the graphs, consistently outperforms the baseline with significance most of the time for both the simulated data and EEG dataset, 
\yitian{with the same order of parameterization.}
Therefore, we conclude that modeling the relation between the features of a time series can lead to improvement in performance.

\begin{figure}
 \centering
 \includegraphics[ width = 0.38\textwidth]{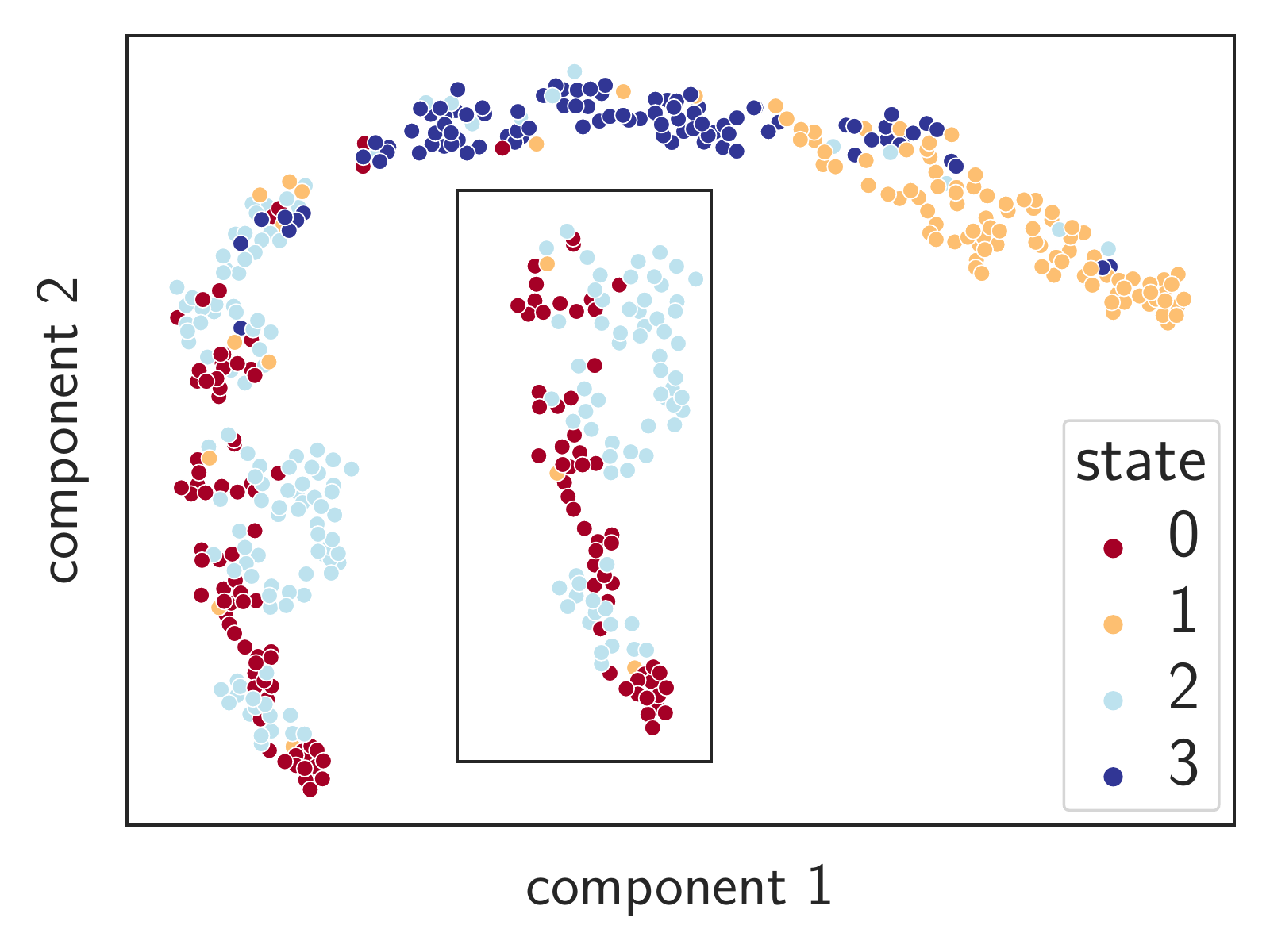}
 \caption*{(a) GraphTNC (ours)}
 \vspace{0.4cm}
 \includegraphics[ width = 0.38\textwidth]{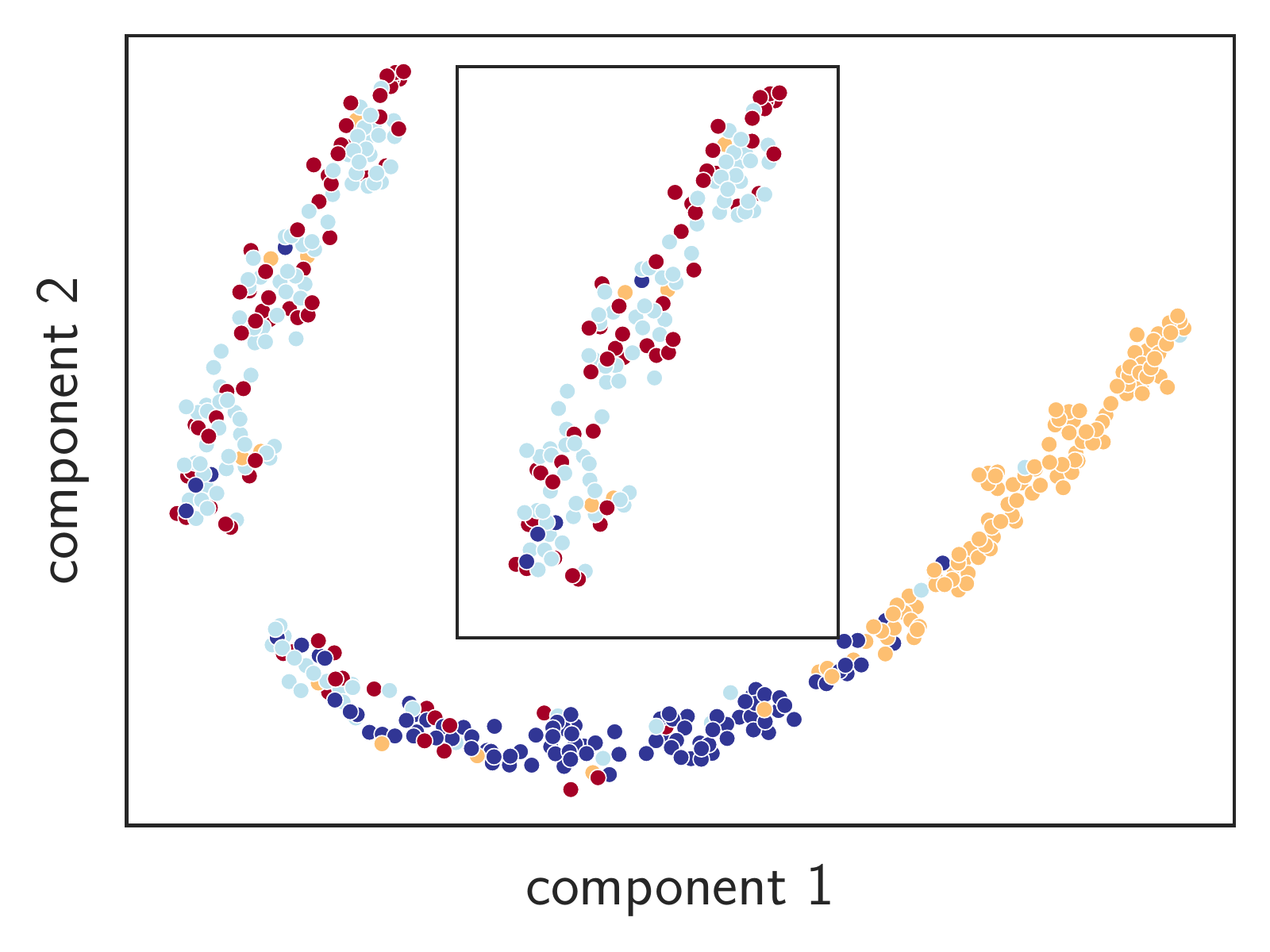}
 \caption*{(b) TNC}
\caption{T-SNE visualization of the 8-dimensional learned encoding $\bz^t$ of windows of simulated dataset with $r=0.1$. The colors represent the underlying state of the windows. The top figure shows the clustering result of GraphTNC (ours), and the bottom figure shows the results for TNC.}
\label{fig:tsne_clustring}
\vspace{-0.05in}
\end{figure}

Second, to further understand how the role of an underlying graph influences the performance, we generate multiple synthetic datasets based on different $r$ parameter values in Equation~\eqref{generate}. A larger $r$ represents that the time series data is more dependent on the graph-defined spacial filtering operation of Eq.~\eqref{generate}. 
We conduct the experiment for $r\in\{0.1, 0.5, 0.9\}$.
We evaluate the model performance by training different models on 10 splits for each $r$ value and report the AUPRC. Our proposed method GraphTNC (in blue) consistently outperforms the time series baseline TNC (in orange). 
This is true for both the accuracy metric and the AUPRC. 
The use of both metrics is important since it accounts for the fact that the label distribution of the synthetic data is not always uniform. AUPRC combines the precision and recall metrics and is known to be a reliable metric for imbalanced datasets.
Besides reporting the mean and standard deviation of the results in Table~\ref{tsgraph}, we also report 95\% confidence intervals in Fig.~\ref{fig:synthetic_results}.
The 95\% confidence intervals are obtained via a non parametric method (bootstrap). We also present a visualisation of the encoding in Fig.~\ref{fig:tsne_clustring} for $r=0.1$. We can see that the GraphTNC representations of are more clearly separated than the TNC representations, especially for states 0 and 2.

\subsubsection{GraphTNC vs non-contrastive learning}

Table~\ref{non-con} displays the classification performance of the representations obtained from the various approaches on the two datasets. The classifier performance of GraphTNC is closer to the supervised model compared to the two other non-contrastive learning methods. They also have similar parameters since the end-to-end learning framework has the same encoder as GraphTNC, followed by a 1-layer FNN. In the EEG dataset, where the state changes infrequently and the graph is static, BYOL and SimSiam can achieve reasonable results. However, when the non-stationarity increases, such as the synthetic data shown in Fig.~\ref{fig:synthetic_data}, the performance drops. Therefore, BYOL and SimSiam, which take neighborhood samples as an augmentation are suitable for more stable time-series scenarios. On the other hand, these two non-contrastive learning approaches need more parameters for training. To conclude,
our proposed GraphTNC is an effective approach for learning representations for non-stationary time series on dynamic graphs. 

\section{Conclusion}

We have introduced an unsupervised learning approach called GraphTNC for data consisting of multivariate time-series on dynamic graphs. Our experimental results for the synthetic and real-world datasets show that the methodology is beneficial when the graphs inform or capture the dynamic relations between features in the signals.

\bibliographystyle{IEEEtran}
\bibliography{ref}

\end{document}